\documentclass[10pt,twocolumn,letterpaper]{article}

\usepackage{iccv}
\usepackage{times}
\usepackage{epsfig}
\usepackage{graphicx}
\usepackage{amsmath}
\usepackage{amssymb}

\usepackage[ruled,linesnumbered]{algorithm2e}
\usepackage{mathtools}
\usepackage{symbols}
\usepackage{siunitx}
\usepackage{bbm}
\usepackage{enumitem}
\usepackage{tabularx}
\usepackage{rotating}
\usepackage{xcolor}
\usepackage[font=small]{caption}
\usepackage{subcaption}
\usepackage{multirow}
\usepackage{booktabs}
\usepackage{url}
\usepackage{makecell}
\usepackage{pifont}
\usepackage[usestackEOL]{stackengine}
\usepackage[symbol]{footmisc}
\stackMath

\definecolor{darkgreen}{rgb}{0.0, 0.5, 0.0}
\definecolor{brickred}{rgb}{0.8, 0.25, 0.33}
\newcommand{\cmark}{\color{darkgreen}\ding{51}}%
\newcommand{\xmark}{\color{brickred}\ding{55}}%

\newcommand\red[1]{\textbf{\textcolor{red}{#1}}}
\newcommand\blue[1]{\textbf{\textcolor{blue}{#1}}}

\usepackage[pagebackref=true,breaklinks=true,letterpaper=true,colorlinks,bookmarks=false]{hyperref}

\iccvfinalcopy 

\def\iccvPaperID{2458} 

\ificcvfinal\fi
\begin{document}

\title{No-Frills Human-Object Interaction Detection: Factorization, Layout Encodings, and Training Techniques}
\author{Tanmay Gupta
\qquad
Alexander Schwing
\qquad
Derek Hoiem\\
University of Illinois Urbana-Champaign\\
{\tt\small \{tgupta6, aschwing, dhoiem\}@illinois.edu \url{http://tanmaygupta.info/no_frills/}}
}

\maketitle
\ificcvfinal\thispagestyle{empty}\fi

\begin{abstract}
    We show that for human-object interaction detection a relatively simple factorized model with appearance and layout encodings constructed from pre-trained object detectors outperforms more sophisticated approaches. Our model includes factors for detection scores, human and object appearance, and coarse (box-pair configuration) and optionally fine-grained layout (human pose). We also develop training techniques that improve learning efficiency by: (1) eliminating a train-inference mismatch; (2) rejecting easy negatives during mini-batch training; and (3) using a ratio of negatives to positives that is two orders of magnitude larger than existing approaches. We conduct a thorough ablation study to understand the importance of different factors and training techniques using the challenging HICO-Det dataset~\cite{chao2018hicodet}. 
\end{abstract}

\section{Introduction}
Human-object interaction (HOI) detection is the task of localizing 
all instances of a predetermined set of human-object interactions. For example, detecting the HOI ``human-row-boat'' refers to localizing a ``human,'' a ``boat,'' and predicting the interaction ``row'' for this human-object pair. Note that an image may contain multiple people rowing boats (or even the same boat), and the same person could simultaneously interact with the same or a different object. For example, a person can simultaneously ``sit on'' and ``row'' a boat while ``wearing'' a backpack.

\begin{figure}[t!]
\vspace{-0.5cm}
    \centering
    \includegraphics[width=0.8\linewidth]{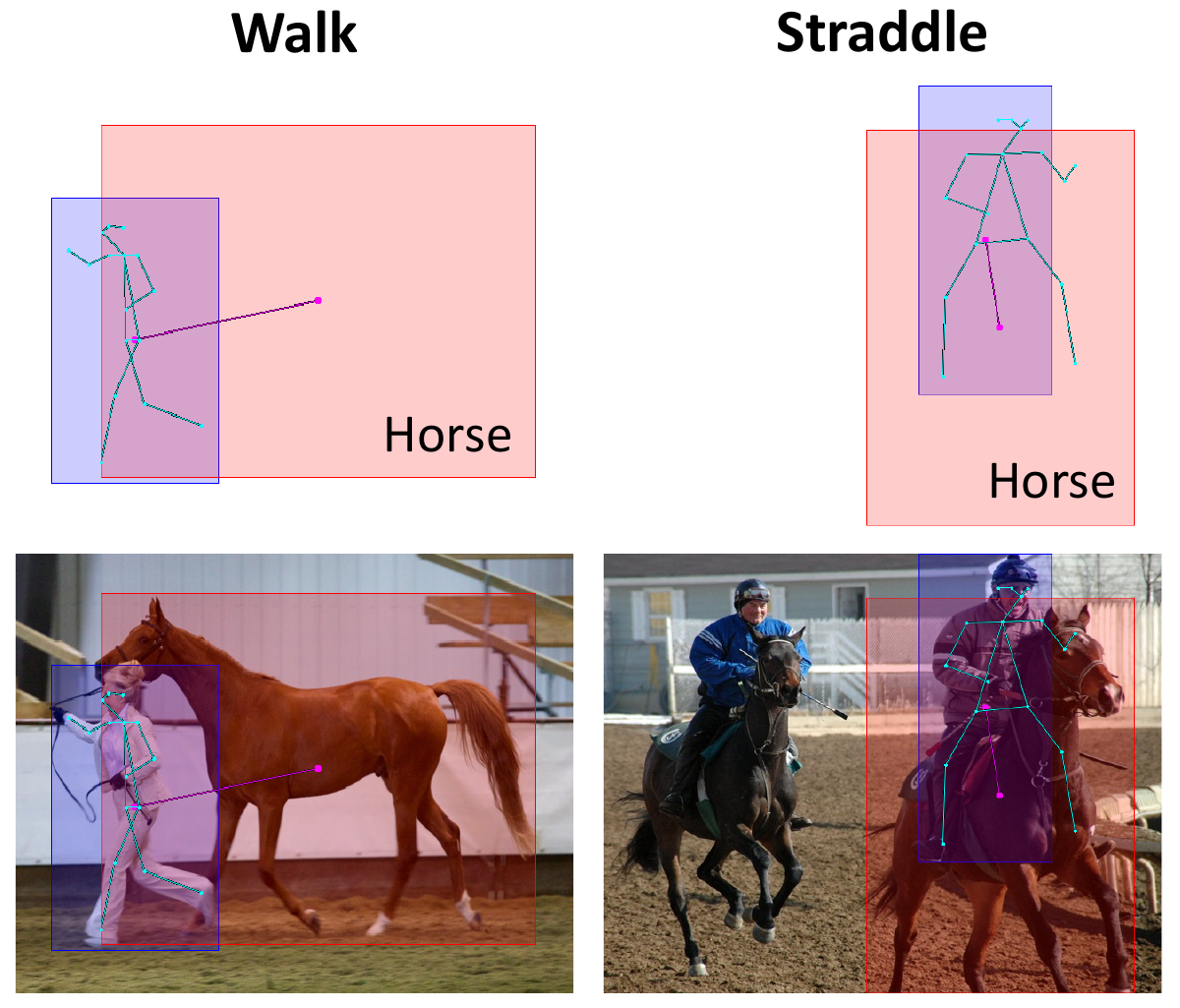}
    \vspace{-0.4cm}
    \caption{\textbf{Outputs of pretrained object and human-pose detectors provide strong cues for predicting interactions}. Top: human and object boxes, object label, and human pose predicted by Faster-RCNN~\cite{ren2015frcnn} and OpenPose~\cite{cao2017realtime} respectively. We encode appearance and layout using these predictions (and Faster-RCNN features) and use a factored model to detect human-object interactions. Bottom: boxes and pose overlaid on the input image.}
    \label{fig:teaser}
    \vspace{-0.5cm}
\end{figure}

Recently, increasingly sophisticated techniques have been proposed for encoding position and appearance for HOI detection. For instance, Chao~\etal~\cite{chao2018hicodet} encode the configuration of human-object box pairs using a CNN operating on a two channel binary image called the \emph{interaction pattern}. Gkioxari~\etal~\cite{gkioxari2018hoi} predict a distribution over target object locations based on human appearance using a mixture density network~\cite{bishop1994mdn}. For encoding appearance, 
approaches range from multitask training of a human-centric branch~\cite{gkioxari2018hoi} alongside  object classification, 
to using an attention mechanism which gathers  contextual information from the image~\cite{Gao2018iCANIA}. 

In this work, we propose a no-frills model for HOI detection. In contrast to sophisticated end-to-end models, we use appearance features from pretrained object detectors, and encode layout using hand-crafted bounding-box coordinate features (optionally human pose keypoints). Our network architecture is also modest, comprising of light-weight multi-layer perceptrons (MLPs) that operate on these appearance and layout features. In spite of these simplifications, our model achieves state-of-the-art performance on the challenging HICO-Det dataset. 

Our gains are due to the choice of factorization, direct encoding and scoring of layout, and improved training techniques. Our model consists of human/object detection terms and an interaction term. The interaction term further consists of human and object appearance, box-configuration, and pose or fine-layout factors. We perform a thorough ablation study to evaluate the effect of each factor. 

\begin{figure*}[t]
\vspace{-0.5cm}
    \centering
    \includegraphics[width=\textwidth]{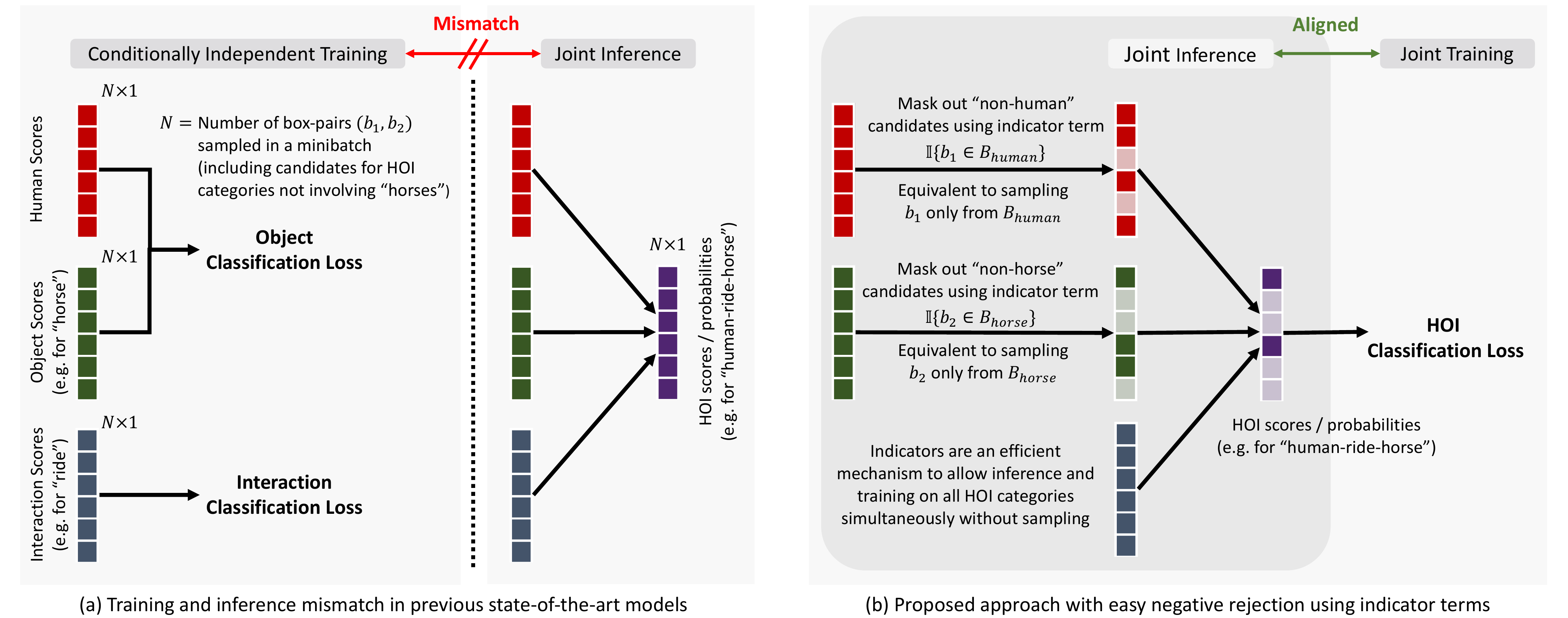}
    \vspace{-0.7cm}
    \caption{\textbf{Proposed training techniques.} The figure shows training and inference on a single HOI class (``human-ride-horse'') for simplicity. \textbf{Eliminating training-inference mismatch:} As shown in (a), existing models~\cite{gkioxari2018hoi,Gao2018iCANIA} often train human/object and interaction branches using object and interaction classification losses. The scores produced by these branches are 
    simply multiplied during testing 
     to produce the final HOI scores. Hence, training does not accurately reflect the test objective. 
     Our model, shown in (b), fixes this mismatch by directly optimizing the combined scores using a multi-label HOI classification loss. \textbf{Rejecting easy negatives:} In a mini-batch, we treat candidate box-pairs for HOI categories other than ``human-ride-horse'' as easy negatives and their probability for HOI category ``human-ride-horse'' is set to $0$ both during training and inference using indicator terms. This is implemented efficiently by applying binary masks to predicted probabilities. Specifically, if a human candidate box $b_1$ does not belong to the set of ``human'' proposals $B_\text{human}$, or if the object candidate box $b_2$ does not belong to the set of ``horse'' proposals $B_\text{horse}$, the mask entry corresponding to $(b_1,b_2)$ and HOI category ``human-ride-horse'' is set to 0.} 
    \label{fig:training_inference}
        \vspace{-0.5cm}
\end{figure*}

In contrast to existing work, which needs to train a CNN~\cite{chao2018hicodet} or a mixture density network~\cite{gkioxari2018hoi} to encode layout, we use hand-crafted absolute and relative position features computed from bounding boxes or human pose keypoints. Our choice is motivated by the observation illustrated in Fig.~\ref{fig:teaser}:  pretrained object and pose detectors provide strong geometric cues for interaction prediction.

We also develop the following training techniques for improving learning efficiency of our factored model:

\noindent
(1) \textbf{Eliminating train-inference mismatch.}~\cite{gkioxari2018hoi,Gao2018iCANIA} learn  detection  and interaction terms via separate detection and interaction  losses. 
During inference, the scores of all factors are simply multiplied 
to get final HOI class probabilities. Instead, we directly optimizing the HOI class probabilities using a multi-label HOI classification loss (Fig.~\ref{fig:training_inference}) (Interaction Loss: $15.89$ mAP \vs HOI Loss: $16.96$ mAP). 
    
\noindent
(2) \textbf{Rejecting easy negatives using indicator terms.} Rejecting easy negatives is beneficial not only during test but also during training because it allows the model to focus on 
learning to score hard negatives. We generate a candidate box-pair $(b_1,b_2)$ using a pre-trained object detector which is then scored by the factor model. If either $b_1$ is not a ``human'' candidate (category $h$) or $b_2$ is not an object candidate $o$, then the factor model should predict a $0$ probability of $(b_1,b_2)$ belonging to HOI category $(h,o,i)$ for \emph{any} interactions $i$. This is achieved by including \emph{indicator} terms in our object detection factors and can be implemented efficiently by applying a mask on predicted probabilities constructed from labels predicted by the object detector (Fig.~\ref{fig:training_inference}) (w/o indicators: $15.93$ mAP \vs w indicators: $16.96$ mAP).

    
\noindent
(3) \textbf{Training with large negative to positive ratio.} We construct training mini-batches by sampling a two orders of magnitude larger number of negative box-pairs per positive pair than related work ($1000$ \vs $<10$). Higher ratios compared to \emph{object} detector training are expected since the number of negative pairs is quadratic in the number of object proposals as opposed to being linear for object detectors (neg. to pos. ratio $10$: $13.40$ mAP \vs $1000$: $16.96$ mAP). 

In summary, our key contributions are: (1) a simple but competitive  model for HOI detection that takes advantage of appearance and layout encodings from a pre-trained object detector (and optionally a pose detector); (2) a comparison of coarse and fine-grained layout encodings; 
and (3) techniques for enhancing learning efficiency of our  model. 

\section{Related Work}
Assessing interactions between humans and objects in images is a challenging problem which has received a considerable amount of attention
~\cite{yao2010mutualcontext,yao2010grouplet,desai2010discriminative,desai2012detecting,delaitre2011learning,maji2011action}. 
 
\noindent \textbf{Human activity recognition} is among the early efforts to analyze human actions in images or videos. Benchmarks such as UCF101~\cite{soomro2012ucf101} and THUMOS~\cite{idrees2017thumos} focus on classifying a video sequence into one of $101$ action categories. While UCF101 uses carefully trimmed videos, 
the THUMOS challenge additionally introduced the task of temporal localization of activities in untrimmed videos. Image action recognition benchmarks such as Stanford 40 Actions~\cite{yao2011human} and PASCAL VOC 2010~\cite{maji2011action} have also been used in the literature. While similar in intent, these action recognition challenges differ from human-object interaction detection in three ways: (1) the tasks are limited to images or videos containing a \emph{single} human-centric action, such as bowling, diving, fencing, \etc; (2) the action classes are \emph{disjoint} and often involve interaction with an object unique to the activity (allowing models to cheat by simply recognizing the object); and (3) spatial localization of neither the person nor the object 
is required. 

\noindent \textbf{Moving from actions to interactions}, Chao \etal~\cite{chao2015hico,chao2018hicodet} introduce the HICO and HICO-DET datasets. 
The HICO dataset consists of images annotated with $600$ human-object interactions with a diverse set of $117$ interactions with $80$ COCO~\cite{lin2014mscoco} object categories. Unlike previous tasks, HOI classification is multi-label in nature since each image may contain multiple humans interacting with the same or different objects. Chao \etal~\cite{chao2018hicodet} extend the HICO dataset with exhaustive bounding box annotations for each of the HOI classes to create HICO-DET. Due to the human-centric nature of the annotation task and the predefined set of objects and interactions, HICO-DET does not suffer from the missing annotation problem (at least to the same extent) that plagues datasets such as Visual Genome~\cite{krishna2017genome} and VRD~\cite{lu2016vrd} that are used for  general visual relationship (object-object interaction) detection. 

In a similar effort, Gupta \etal~\cite{gupta2015vsrl} augment the COCO dataset~\cite{lin2014mscoco}, annotating people (agents) with one of $26$ action labels along with location and labels of objects fulfilling 
 semantic roles for the action. 
Moreover, for semantic role labeling (SRL), 
Yatskar \etal~\cite{yatskar2016situation} create an image dataset for situation recognition, which 
subsumes recognition of activity, participating objects and their roles. 

In this work, we choose HICO-DET as a test bed 
due to its large, diverse, and exhaustive HOI annotations which allows for an accurate and meaningful evaluation. 


\noindent \textbf{Existing models for HOI detection.}  Chao \etal~\cite{chao2018hicodet} propose HO-RCNN, a $3$-stream architecture with one stream each for a human candidate, an object candidate, and a geometric encoding of the pair of boxes using the proposed \emph{interaction pattern}. Each stream produces scores for every possible object-interaction category ($600$ for HICO-DET). The $3$ sets of scores are combined using late-fusion to get the final prediction. Note that this approach treats ``ride bicycle'' and ``ride horse'' as independent visual entities and does not use the knowledge of ``ride'' being a common component. In contrast, our approach exploits this compositionality to learn shared visual appearance and geometric representations (\eg, ``ride'' typically involves a human box above an object box). In other words, weight sharing between different HOI classes in our factored model makes it more data efficient than~\cite{chao2018hicodet} which predicts scores for $600$ HOI categories using independent weights in the last $600$-way fully connected layer in each of the 3 streams. 

Gkioxari \etal~\cite{gkioxari2018hoi} propose \emph{InteractNet}, which takes a multitask learning~\cite{Caruana1998Multitask} perspective. 
The idea is to augment the Faster-RCNN~\cite{ren2015frcnn} object detection framework with a human-centric branch and an interaction branch that are trained jointly alongside the original object recognition branch. To incorporate geometric cues, a Mixture Density Network (MDN)~\cite{bishop1994mdn} is used to produce parameters of the object location distribution given the human appearance. This distribution is used to score candidate objects for a given human box. The model is trained using an object classification loss for the object branch, interaction classification losses for the human-centric action classification branch and the optional interaction branch, and a smooth L1 loss between the ground truth box-pair encoding and mean predicted by the localization MDN. During inference, predictions from these branches are fused heuristically. In addition to differences in the details of factorization, and appearance and layout encodings used in our model, we introduce training techniques for enhancing learning efficiency of similar factored models for this task.  We optimize the final HOI score obtained after fusing the individual factor scores. We also encode more directly a  box-pair layout using absolute and relative bounding box features which are then scored using a dedicated factor. 

Gao \etal~\cite{Gao2018iCANIA} follow an approach similar to~\cite{gkioxari2018hoi} but introduce an attention mechanism that augments human and object appearance with contextual information from the image. An attention map is computed using cues derived from the human/object appearance encoding and the context is computed as an attention weighted average of convolution features. The model is trained using an interaction classification loss. In contrast, the only sources of contextual information in our model are the ROI pooled region features from the object detector. Adding a similar attention mechanism may further improve performance. 

\section{No-Frills HOI Detection}
In the following, we first present an overview of the proposed  model, followed by details of different factors and our training strategy. 

\newcommand\mycommfont[1]{\footnotesize\ttfamily\textcolor{red}{#1}}
\SetCommentSty{mycommfont}

\makeatletter
\newcommand{\removelatexerror}{\let\@latex@error\@gobble}
\makeatother

\begin{figure}[t]
\begin{minipage}{\linewidth}
\begingroup
\removelatexerror
\vspace{-0.75em}
\begin{algorithm}[H] 
\footnotesize
\caption{Inference on a single image}
\label{alg:inference}
\SetKwInOut{Input}{Input}
\SetKwInOut{Output}{Output}

\Input{
    Image $x$, \\
    Set of objects ($\mathcal{O}$), interactions ($\mathcal{I}$), and \\ 
    \hspace{0.5cm}HOI ($\mathcal{H} \subseteq \{h\} \times \mathcal{O} \times \mathcal{I}$) classes of interest, \\
    Pretrained object (Faster-RCNN) and \\
    \hspace{0.5cm}human-pose (OpenPose) detectors
}
\tcp{\textbf{Stage 1: Create a set of box candidates for each object (including human)}}
Run Faster-RCNN on $x$ $\forall \; o \in \mathcal{O}$ to get $300$ region proposals ($R_o$) \\\hspace{0.5cm}with ROI appearance features and detection probabilities \\

\ForEach{$o \in \mathcal{O}$}{
    Construct $B_o=\{b\in R_o \; \textrm{such that} \;$\\
    \hspace{2.1cm}$b \; \textrm{survives NMS (threshold 0.3) and}$ \\
    \hspace{2.1cm}$P_\text{det}(l_\text{det}=o|b,x) > 0.01\}$ \\
    Update $B_o$ to keep at most 10 highest ranking detections. 
}

Run OpenPose on $x$ to get skeletal-keypoints $k(b) \; \forall \; b \in B_h$ \\
\tcp{\textbf{Stage 2: Score candidate pairs using the proposed factored model}}
\ForEach{$(h,o,i) \in \mathcal{H}$}{
    \ForEach{$(b_1,b_2) \in B_h\times B_o$}{
        Compute box configuration features using $(b_1,b_2)$ \\
        Compute pose features using $(k(b_1),b_1,b_2)$ \\
        Compute $P(y_{(h,o,i)}=1|b_1,b_2,x)$ \\ 
        \hspace{0.5cm}using equations \ref{eq:prob_hoi}, ~\ref{eq:prob_det}, and \ref{eq:prob_i}
    }
    \Output{Ranked list of $(b_h,b_o) \in B_h \times B_o$ as detections for class $(h,o,i)$ with probabilities. For any $o'\neq o$ probability of $(b_h,b_{o'})$ belonging to class $(h,o,i)$ is predicted as 0.}
}
\tcp{\textbf{Steps 10-17 are implemented with a single forward pass on a mini-batch of precomputed features}}
\end{algorithm} 
\endgroup
\end{minipage}
\vspace{-0.5cm}
\end{figure}

\subsection{Overview}
Given an image $x$ and a set of object-interaction categories of interest, human-object interaction (HOI) detection is the task of localizing all human-object pairs participating in one of the said interactions. The combinatorial search over human and object bounding-box locations and scales, as well as object labels, $\mathcal{O}$, and interaction  labels, $\mathcal{I}$,  makes both learning and inference challenging. To deal with this complexity, we decompose  inference into two stages (\algref{alg:inference}). In the \textbf{first stage}, object category specific bounding box candidates $B_o \; \forall \; o \in \mathcal{O}$ are selected using a \textit{pre-trained} object detector such as Faster-RCNN (using non-maximum suppression and thresholding on class probabilities). For each HOI category, \ie, for each triplet $(h,o,i)\in\mathcal{H}$, a set of candidate human-object box-pairs is constructed by pairing every human box candidate $b_h \in B_h$ with every object box candidate $b_o \in B_o$. In the \textbf{second stage}, a factored model is used to score and rank candidate box-pairs $(b_h,b_o)\in B_h\times B_o$ for each HOI category. Our factor graph consists of human and object appearance, box-pair configuration (coarse layout) and human-pose (fine-grained layout) factors. The factors operate on appearance and layout encodings constructed from outputs of pretrained object and human-pose detectors. The model is parameterized to share representations and computation across different object and interaction categories to efficiently score candidate box-pairs for all HOI categories of interest in a single forward pass. 

\subsection{Factored Model}
\vspace{-0.2cm}
For an image $x$, given a human-object candidate box pair $(b_1,b_2)$, human pose keypoints $k(b_1)$ detected inside $b_1$ (if any), and the set of box candidates for each object category, the factored model computes the probability of occurrence of human-object interaction $(h,o,i)$ in $(b_1,b_2)$ as follows:
\begin{eqnarray} 
\label{eq:prob_hoi}
    &&\hspace{-0.5cm}P(y_{(h,o,i)}=1|b_1,b_2,x,k(b_1),B_h,B_o)\\
    &&\hspace{-0.5cm}=P(y_h=1,y_o=1,y_i=1|b_1,b_2,x,k(b_1),B_h,B_o) \nonumber \\
    &&\hspace{-0.5cm}=\Longunderstack[l]{
        P(y_h=1|b_1,x,B_h) \cdot P(y_o=1|b_2,x,B_o) \; \cdot \\
        P(y_i=1|b_1,b_2,k(b_1),x).}\nonumber
\end{eqnarray}

Here, $y_h\in\{0,1\}$ is a random variable denoting if $b_1$ is labeled as a human, $y_o\in\{0,1\}$ denotes if $b_2$ is labeled as object category $o$, and $y_i\in\{0,1\}$ denotes if the interaction assigned to the box-pair is $i$. The above factorization assumes that human and object class labels depend only on the individual boxes, the image, and the set of bounding-box candidates for the respective classes, while the interaction label depends on the box-pair, pose, 
and the image. $B_h$ and $B_o$ are used in the detection terms to compute the indicator functions for easy negative rejection. For brevity, we refer to the left hand side of Eq.~\eqref{eq:prob_hoi} as $P(y_{(h,o,i)}=1|b_1,b_2,x)$. We now describe the terms in detail.

\vspace{-0.2cm}
\subsubsection{Detector Terms}
\vspace{-0.2cm}
The first two terms in Eq.~\eqref{eq:prob_hoi} are modeled using the set of candidate bounding boxes for each object class and classification probabilities produced by a pretrained object detector. For any object category $o \in \mathcal{O}$ (including $h$), the detector term is computed via 
\begin{equation} \label{eq:prob_det}
    P(y_o=1|b,x,B_o) = \mathbbm{1}(b\in B_o) \cdot P_\text{det}(l_\text{det}=o|b,x),
\end{equation}
where the $P_\text{det}$ term corresponds to the probability of assigning object class $o$ to region $b$ in image $x$ by the object detector. The indicator function checks if $b$ belongs to $B_o$, the set of candidate bounding boxes for $o$, and sets the probability to $0$ otherwise. Thus, easy negatives for class $(h,o,i)$, \ie, pairs $(b_1,b_2)$ where either $b_1 \notin B_h$ or $b_2 \notin B_o$, are assigned $0$ probability. Easy negative rejection is not only beneficial during test but also during training since model capacity is not wasted on learning to predict a low probability of class $(h,o,i)$ for box-pairs that belong to any of the classes in the set $\{(h,o',i') \; | \; o'\in \mathcal{O}\setminus\{o\},\; i'\in \mathcal{I}\}$. 

\begin{figure*}[t]
    \vspace{-0.5cm}
    \centering
    \includegraphics[width=\textwidth]{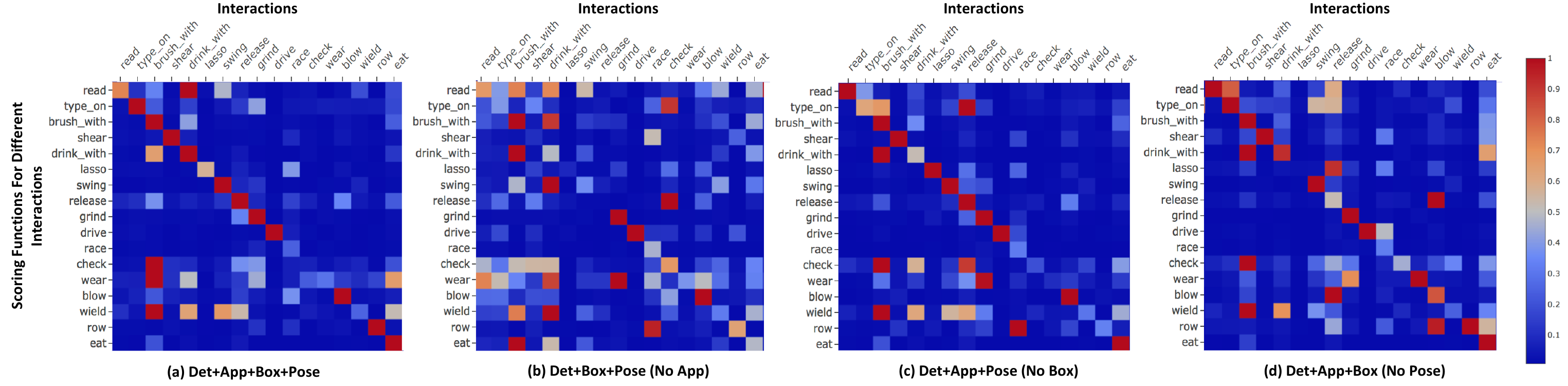}
    \vspace{-0.8cm}
    \caption{\textbf{Interaction confusions.} Element $(m,n)$ in each heatmap visualizes ${P(y_{i_m}=1|b_1,b_2,k(b_1),o,x)}$, the probability of interaction $i_m \in \mathcal{I}$ for box-pair $(b_1,b_2)$, averaged across all box pairs with ground truth interaction $i_n \in \mathcal{I}$. Each row $m$ is independently normalized and exponentiated to highlight the interactions most confused with interaction $i_m$.} 
    \label{fig:interact_heatmaps}
        \vspace{-0.5cm}
\end{figure*}

\vspace{-0.2cm}
\subsubsection{Interaction Term}
\vspace{-0.2cm}
The interaction term refers to the probability of entities in $b_1$ and $b_2$ engaging in interaction $i\in \mathcal{I}$. To utilize appearance and layout information, the interaction term ${P(y_i=1|b_1,b_2,k(b_1),x)}$ is further factorized as follows:
\begin{equation} \label{eq:prob_i}
\begin{aligned}
    \sigma(\Longunderstack[l]{
    \phi_\text{human}(i|b_1,x) + 
    \phi_\text{object}(i|b_2,x) \; + \\
    \phi_\text{boxes}(i|b_1,b_2) +
    \phi_\text{pose}(i|b_1,b_2,k(b_1))),}
\end{aligned}
\end{equation}
where $\sigma$ is the Sigmoid function and each $\phi$ is a learnable deep net factor. We now describe each of these factors along with the network architecture and appearance and layout encodings the factors operate with: 

\noindent \textbf{Appearance.} Factors $\phi_\text{human}$ and $\phi_\text{object}$ predict the interaction that the human and the object are engaged in, based on visual appearance alone. The appearance of a box in an image is encoded using Faster-RCNN~\cite{ren2015frcnn} (Resnet-152 backbone) average pooled \textit{fc7} features extracted from the RoI. By design, this representation captures context in addition to content within the box. The $2048$ dimensional \textit{fc7} features are fed into a multi-layer perceptron (MLP) with a single $2048$ dimensional hidden layer with Batch Normalization~\cite{ioffe2015batchnorm} and ReLU~\cite{nair2010relu}. The output layer has $117$ neurons, one per interaction category in $\mathcal{I}$. 

\noindent \textbf{Box Configuration.} Object label and the absolute and relative positions and scales of the human and object boxes are often indicative of the interaction, even without  the appearance (\eg, a human box above and overlapping with a `horse' box strongly suggests a `riding' interaction). $\phi_\text{boxes}$ captures this intuition by predicting a score for each interaction given an encoding of the bounding boxes and the object label. 
The bounding boxes are represented using a 21-dimensional feature vector. We encode the \textit{absolute position and scale} of both the human and object boxes using box width, height, center position, aspect ratio, and area. We also encode \textit{relative configuration} of the human and object boxes using relative position of their centers, ratio of box areas and their intersection over union. 

We also experiment with conditioning on the object label encoded as a $|\mathcal{O}|$ ($=80$) dimensional one hot vector to allow the model to learn that only certain interactions are feasible for a given object. For example, it is possible to ``clean'' or ``eat at'' a ``dinning table'' ($o$) but not to ``drive'' or ``greet'' it. The object label encoding is concatenated with 21-dimensional position features and their log absolute values, and passed through an MLP with 2 hidden layers, 112 ($=2 \times 21 + 80$) 
dimensional each (same as the input feature dimension), with Batch Normalization and ReLU. In practice, conditioning on $o$ did not affect results significantly and hence is not included in Eq.~\ref{eq:prob_hoi} and Eq.~\ref{eq:prob_i} for simplicity.

\noindent \textbf{Human Pose.} We supplement the coarse layout encoded by bounding boxes with more fine-grained layout information provided by human pose keypoints. We use OpenPose~\cite{cao2017realtime,wei2016cpm,simon2017hand} to detect 18 keypoints for each person in the image. A human candidate box is assigned a keypoint-skeleton if the smallest bounding box around the keypoints has $70\%$ or more of its area inside the human box. Similar to box features, we encode both absolute human pose and the relative location with respect to the object candidate box. The \emph{absolute pose features} ($18 \times 3=54$) consist of keypoint coordinates normalized to the human bounding box frame of reference and confidence of each keypoint predicted by OpenPose. The \emph{relative pose features} ($18 \times 5=90$) consist of offset of the top left and bottom right corners of the object box relative to each keypoint and keypoint confidences. The absolute and relative pose features and their log values are concatenated along with a one-hot object label encoding before being passed through $\phi_\text{pose}$. $\phi_\text{pose}$ is also an MLP with 2 hidden layers with $368$ ($=2\times (54 + 90) + 80$) neurons each. Both hidden layers are equipped with Batch Normalization and ReLU. The output layer 
has 117 neurons.

\noindent\textbf{Each factor eliminates some interaction confusions.} Heatmaps in Fig.~\ref{fig:interact_heatmaps} show the interactions that are confused by models with all factors and models with one factor missing at a time. Comparing heatmap \textbf{b} with \textbf{a} shows the role of the appearance factor in reducing confusion between interactions. For instance, without \emph{App}, ``eat'' is confused with ``brush with'' and ``drink with,'' but not in the final model. Similarly, \textbf{c} and \textbf{d} can be compared with \textbf{a} for the effects of \emph{Box} and \emph{Pose} factors respectively.

\subsection{Training}\label{sec:training}
Since more than one HOI label might be assigned to a pair of boxes, the model is trained in a fully supervised fashion using the multi-label binary cross-entropy loss.  For each image in the training set, candidate boxes for each HOI category ($B_h\times B_o$ for class $(h,o,i)$) are assigned binary labels based on whether both the human and object boxes in the pair have an intersection-over-union (IoU) greater than $0.5$ with a ground truth box-pair of the corresponding HOI category. During training, the $j^{\text{th}}$ sample in a mini-batch consists of a box pair $(b_1^j,b_2^j)$, HOI category $l_j \in \mathcal{H}$ for which the box pair is a candidate ($(b_1,b_2) \in B_h\times B_o$ are considered candidates for HOI class $(h,o,i)$), binary label $y^j$ to indicate match (or not) with a ground truth box pair of class $l_j$, detection scores for human and object category corresponding to class $l_j$, and input features for each factor $\phi$. Pairs of boxes which are candidates for more than one HOI category are treated as multiple samples during training. Since the number of candidate pairs per image is 3 orders of magnitude (typically $>1000$) larger than the number of positive samples (typically $<3$), random sampling would leave most mini-batches with no positives. We therefore select all positive samples per image and  randomly sample 1000 negatives per positive.  Given a mini-batch of size $N$ constructed from a single image $x$, the loss is
\begin{equation}~\label{eq:ch4_training_loss}
\hspace{-0.5cm}    \begin{aligned}
    \mathcal{L}_{\text{mini-batch}} = &\frac{1}{N|\mathcal{H}|}
    \sum_{j=1}^{N}
    \sum_{l\in \mathcal{H}}
    \mathbbm{1}(l=l^j) \cdot \textrm{BCE}(y^j,p_l^j),
    \end{aligned}
\end{equation}
where $\textrm{BCE}(y,p)$ 
is the binary cross entropy loss and ${p_l^j=P(y_l=1|b_1^j,b_2^j,x)}$ is the probability of HOI class $l$ computed for the $j^\text{th}$ sample 
using Eq.~\ref{eq:prob_hoi}. In our experiments, we only learn parameters of the interaction term (\ie, MLPs used to compute factors $\phi_\text{human}$, $\phi_\text{box},$ and $\phi_\text{pose}$).

\renewcommand{\arraystretch}{1.1}
\setlength{\tabcolsep}{6pt}
\begin{table}[t]
\vspace{-0.5cm}
    \footnotesize
    \centering
    \addtolength{\tabcolsep}{0pt}
    \begin{tabular}{l|c|c|c}
        \textbf{Models} & \textbf{Full}  & \textbf{Rare}  & \textbf{Non-Rare} \\
        \hline
        HO-RCNN~\cite{chao2018hicodet}   & 7.81  & 5.37  & 8.54     \\
        VSRL~\cite{gupta2015vsrl} (impl. by ~\cite{gkioxari2018hoi})       & 9.09  & 7.02  & 9.71     \\
        InteractNet~\cite{gkioxari2018hoi} & 9.94  & 7.16  & 10.77    \\
        GPNN~\cite{qi2018graphparsingnn} & 13.11  & 9.34  & 14.23    \\
        iCAN~\cite{Gao2018iCANIA} & 14.84  & 10.45  & 16.15    \\ 
        \hline
        Det                    & 8.32  & 6.84  & 8.76  \\
        Det + Box              & 12.54 & 10.40 & 13.18 \\
        Det + Human App        & 11.12 & 8.82  & 11.80 \\
        Det + Object App       & 11.05 & 7.41  & 12.13 \\
        Det + App              & 15.74 & 11.35 & 17.05 \\
        Det + Human App + Box  & 15.63 & \red{12.45} & 16.58 \\
        Det + Object App + Box & 15.68 & 10.47 & 17.24 \\
        Det + App + Box        & \blue{16.96} & 11.95 & \blue{18.46} \\
        \hline
        Det + Pose             & 11.09 & 8.04  & 12.00 \\
        Det + Box + Pose       & 14.49 & 11.86 & 15.27 \\
        Det + App + Pose       & 15.50 & 10.14 & 17.10 \\
        Det + App + Box + Pose & \red{17.18} & \blue{12.17} & \red{18.68} \\
    \end{tabular}
    \vspace{-0.3cm}
    \caption{\textbf{Results on HICO-Det test set.} Det, Box, App, and Pose correspond to object detector terms, appearance, box configuration, and pose factors respectively. Each row was both trained and evaluated with specified factors. \emph{Full} (all $600$ classes), \emph{Rare} (classes with $<10$ training instances), and \emph{Non-Rare} (rest) denote different subsets of HOI classes. \red{Best} and \blue{second best} numbers are highlighted in color.}
    \label{tbl:ch4_factor_ablation}
    \vspace{-0.5cm}
\end{table}
\renewcommand{\arraystretch}{1}

\section{Experiments}
\noindent\textbf{Dataset.} HICO-Det~\cite{chao2018hicodet} and V-COCO~\cite{gupta2015vsrl} datasets are commonly used for evaluating HOI detection models. V-COCO is primarily used for legacy reasons since the early HICO~\cite{chao2015hico} dataset only had image-level annotations. HICO-Det was created to extend HICO with bounding box annotations specifically for the HOI detection task. HICO-Det is both larger and more diverse than V-COCO. While HICO-Det consists of $47,776$ images annotated with $117$ interactions with $80$ objects resulting in a total of $600$ HOI categories, V-COCO only has $26$ interactions with a training set $1/12$ the size of HICO-Det's. Exhaustive annotations for each HOI category also make HICO-Det more suitable for $AP$ based evaluation than VRD~\cite{lu2016vrd} which suffers from missing annotations. VRD also contains ``human'' as one among many subjects which makes evaluation of the impact of fine-grained human pose less reliable due to a small sample size. Hence, HICO-Det is best for evaluating our contributions. 

HICO-Det contains $38,118$ training and $9,658$ test images annotated with $600$ HOI categories. We use an $80$-$20$ split of the training images to generate our actual training and validation sets. 
HOI categories consist of $80$ object categories (same as COCO classes) and $117$ interactions. Each image contains on average $1.67$ HOI detections.

\renewcommand{\arraystretch}{1.1}
\setlength{\tabcolsep}{2.7pt}
\begin{table}[t]
\vspace{-0.5cm}
    \footnotesize
    \centering
    \addtolength{\tabcolsep}{0pt}
    \footnotesize
    \begin{tabular}{c|c|c|c|c}
        \textbf{Neg./Pos.} & \textbf{Indicators} & \textbf{HOI Loss} & \textbf{Interaction Loss} & \textbf{mAP} \\
        \hline
        10 & \cmark & \cmark & \xmark & 13.40 \\
        50 & \cmark & \cmark & \xmark & 15.51 \\
        100 & \cmark & \cmark & \xmark & 16.30 \\
        \red{500} & \cmark & \cmark & \xmark & \red{17.06} \\
        \blue{1000} & \cmark & \cmark & \xmark & \blue{16.96} \\
        1500 & \cmark & \cmark & \xmark & 16.62 \\
        \hline
        1000 & \xmark & \cmark & \xmark & 15.93 \\
        1000 & \cmark & \xmark & \cmark & 15.89 \\
    \end{tabular}
    \vspace{-0.3cm}
    \caption{\textbf{Training techniques evaluated using Det + App + Box model.} The results highlight the importance of: (1) large \emph{negative to positive ratio} in mini-batches; (2) using \emph{indicators} during training to only learn to rank candidates selected specifically for a given HOI category instead of all detection pairs; (3) directly optimizing the \emph{HOI classification} loss instead of training with an \emph{interaction classification} loss and then combining with object detector scores heuristically. \red{Best} and \blue{second best} numbers are shown in color.}
    \label{tbl:ch4_design_choices}
    \vspace{-0.5cm}
\end{table}
\renewcommand{\arraystretch}{1}

\begin{figure*}[t!]
    \vspace{-0.5cm}
    \centering
    \includegraphics[width=0.9\linewidth]{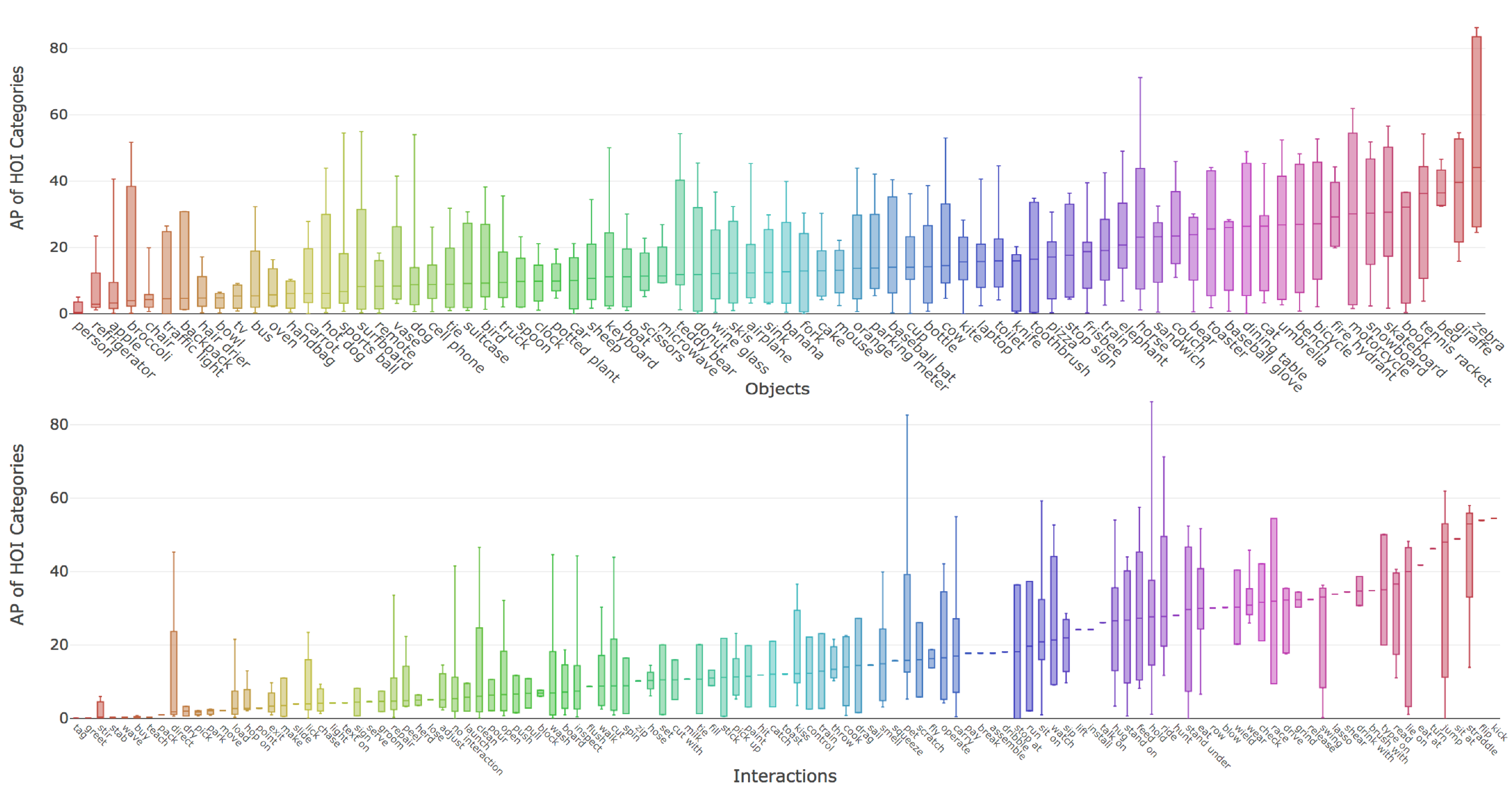}
        \vspace{-0.5cm}
    \caption{\textbf{Spread of performance} (range and quartiles) across interactions with the same object (top) and across objects for a given interaction (bottom). The horizontal axis is sorted by median AP.}
    \label{fig:spread_of_perf}
    \vspace{-0.5cm}
\end{figure*}

\begin{figure*}[t]
    \vspace{-0.5cm}
    \centering
    \includegraphics[width=0.9\textwidth]{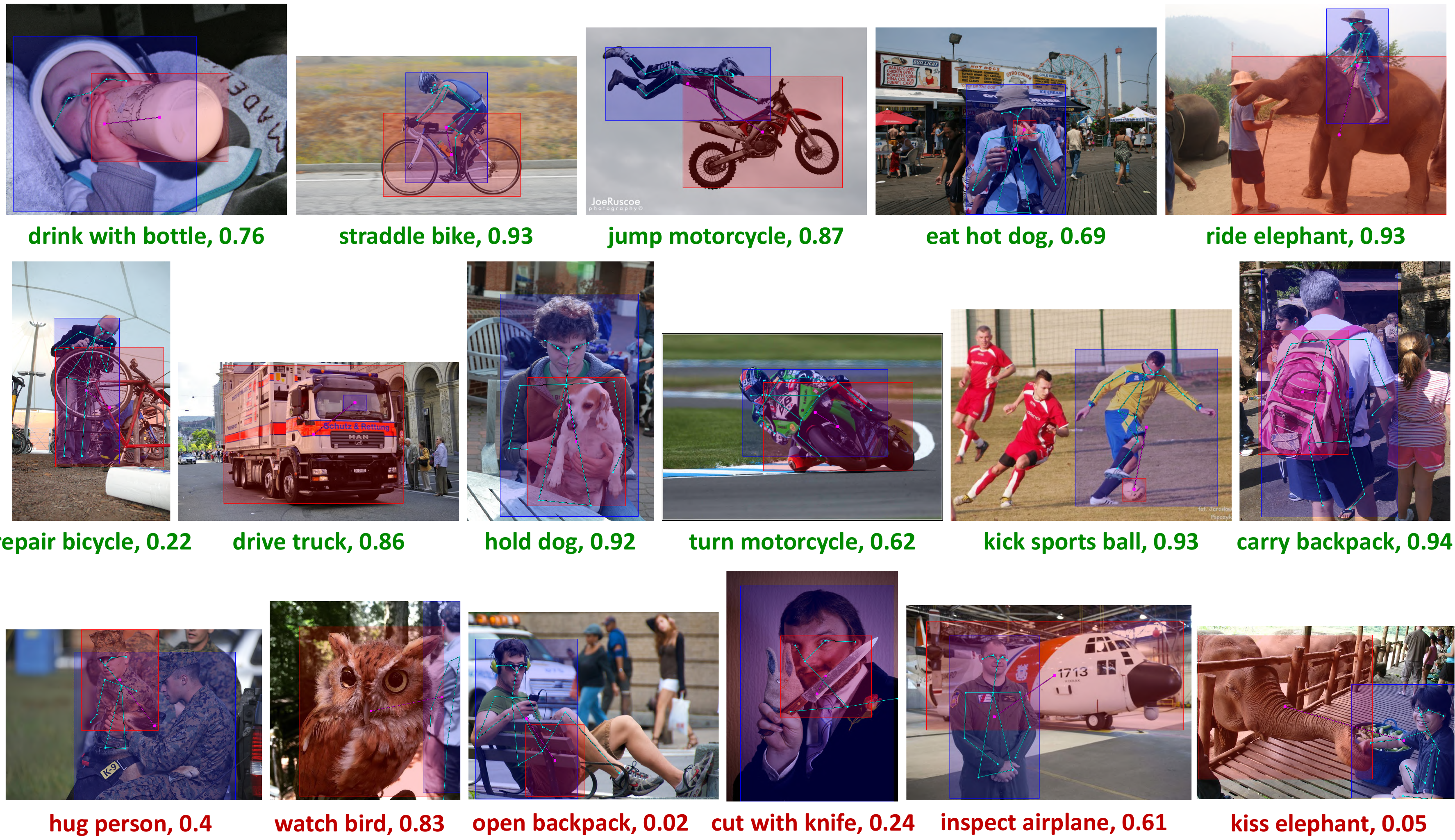}
    \vspace{-0.3cm}
    \caption{\textbf{Qualitative results} showing top ranking \textbf{\color{darkgreen}true} and \textbf{\color{brickred}false} positives for different HOI categories with predicted probability. The blue and red boxes correspond to human and objects detected by a pretrained Faster-RCNN detector respectively. Pose skeleton consists of 18 keypoints predicted by the pretrained OpenPose detector and assigned to the human box.}
    \label{fig:ch4_qual_results}
    \vspace{-0.5cm}
\end{figure*}

In addition to comparing to state-of-the-art, our experiments include factor ablation study (Tab.~\ref{tbl:ch4_factor_ablation}), impact of the proposed training techniques (Tab.~\ref{tbl:ch4_design_choices}), visualization of performance distribution across object and interaction categories (Fig.~\ref{fig:spread_of_perf}), and examples of top ranking detections and failure cases (Fig.~\ref{fig:ch4_qual_results}).




\subsection{Comparison to State-of-the-art}  
Tab.~\ref{tbl:ch4_factor_ablation} shows that our final models \emph{Det+App+Box} and \emph{Det+App+Box+Pose} outperform existing approaches. We now highlight the key strengths of our approach in comparison to existing models.

\noindent
\textbf{Appearance does not need to be relearned.} All existing approaches learn a task specific appearance encoding by either fine-tuning appearance encoding branches~\cite{gkioxari2018hoi,Gao2018iCANIA,gupta2015vsrl} or training a CNN from scratch~\cite{chao2018hicodet}. We only use ROI pooled features from Faster-RCNN pretrained on MS-COCO~\cite{lin2014mscoco}. 

\noindent
\textbf{Layout is directly encoded and scored.} We directly encode layout using absolute and relative position features which are scored using $\phi_\text{box}$ (an MLP). Our formulation is easier to learn than \emph{InteractNet} which predicts a distribution over target object locations using human appearance features alone. The explicit representation also makes our layout terms more efficient to learn than \emph{HO-RCNN} which needs to learn to encode layout (\emph{interaction pattern}) using a CNN. 

\noindent
\textbf{Weight sharing for learning efficiency.} Weight sharing in our factored model (also in \emph{InteractNet} and \emph{iCAN}) makes it more data efficient than \emph{HO-RCNN} which predicts scores for 600 HOI categories using independent weights in the last 600-way fully connected layer. In other words, \emph{HO-RCNN} treats ``ride-bike'' and ``ride-horse'' as independent visual entities and does not benefit from the knowledge of ``ride'' being a common component.

\noindent
\textbf{ROI pooling for context.} \emph{iCAN} follows an approach similar to \emph{InteractNet} but augments region appearance features with contextual features computed using an attention mechanism. While our model demonstrates strong performance with only ROI pooled detector features as the source of contextual information, we may further benefit from an attention mechanism similar to \emph{iCAN}. \emph{GPNN} also attempts to benefit from global context reasoning through message passing over an inferred graph. While in theory, such an approach \emph{jointly} infers all HOI detections in an image (as opposed to making predictions for one candidate box-pair at a time), the advantages of this approach over simpler fixed graph approaches like our factor model and $iCAN$ 
remains to be demonstrated.

Our model also benefits from improved training techniques which are discussed next.

\subsection{Training Techniques}\label{sec:design_choices}
Tab.~\ref{tbl:ch4_design_choices} shows the effect of proposed training techniques.

\noindent\textbf{Training with large negative to positive ratio.} Increasing the ratio of negative to positive box-pairs in a mini-batch during training leads to a dramatic increase in performance (neg. to pos. ratio $10$: $13.40$ mAP \vs $1000$: $16.96$ mAP). Note that related work~\cite{chao2018hicodet,gkioxari2018hoi} uses low ratios (typically $< 10$), similar to those used for training object detectors. For HOI detection, since the number of negative pairs is quadratic in the number of object proposals as opposed to linear for object detectors, higher ratios are expected.

\noindent\textbf{Eliminating train-inference mismatch.} Training the model using interaction classification loss on the probabilities predicted by the interaction term, as done in~\cite{gkioxari2018hoi,Gao2018iCANIA}, is suboptimal in comparison to training using HOI classification loss ($15.89$ \vs $16.96$ mAP) even though the same set of parameters are optimized by both losses. This is because the latter calibrates the interaction term relative to the detection terms. A similar approach is used in~\cite{chao2018hicodet} but without the strong weight sharing assumptions of our factor model. 

\noindent\textbf{Rejecting easy negatives.} To allow the model to focus on learning to rank hard negatives correctly, we introduce indicator terms in our factor model. The indicator functions ensure that the factor model predicts zero probability of an HOI category $(h,o,i)$ for box-pair $(b_1,b_2)$ if $b_1\not\in B_h$ or $b_2\not\in B_o$. Tab.~\ref{tbl:ch4_design_choices} shows that removing the indicator terms during training causes a drop in mAP from $16.96$ to $15.93$ (indicators still used during inference).

\subsection{Factor Ablation Study} 
To identify the role of different sources of appearance and spatial information in our model, we train models with subsets of available factors. 

The role of individual factors can be assessed by comparing \emph{Det}, \emph{Det+Box}, \emph{Det+App}, and \emph{Det+Pose}. Note that appearance terms lead to largest gains over \emph{Det} followed by \emph{Box} and \emph{Pose}. We further analyze the contribution of human and object appearance towards predicting interactions. Interestingly, while \emph{Det+Human App} and \emph{Det+Object App} perform comparably ($11.12$ and $11.05$), the combination outperforms either of them with an mAP of $15.74$, showing that the human and object appearance provide  complementary information. Note that an mAP of $11.12$ ($=\textrm{max}(11.12,11.05)$) or less would indicate completely redundant or noisy signals. A similar sense of complementary information can be observed in Tab.~\ref{tbl:ch4_factor_ablation} for \emph{App-Box}, \emph{App-Pose}, and \emph{Box-Pose} pairs. 

While \emph{Det+Box+Pose} improves over \emph{Det+Box}, \emph{Det+App+Pose} and \emph{Det+App} perform comparably. Similarly, \emph{Det+App+Box+Pose} only slightly improves the performance of \emph{Det+App+Box}. This suggests that while it is useful to encode fine-grained layout in addition to coarse layout, human appearance encoded via object detectors  already  captures human pose  to some extent. 

Another way of understanding the role of factors is to consider the drop in performance when a particular factor is removed from the final model. Relative to \emph{Det+App+Box+Pose}, performance drops are $2.69$, $1.68$, and $0.22$ mAP for \emph{App}, \emph{Box} and \emph{Pose} factors respectively.

\subsection{Performance Distribution}
Fig.~\ref{fig:spread_of_perf} visualizes the distribution of performance of our model across interactions with a given object and across objects for a given interaction. The figure shows that for most objects certain interactions are much  easier to detect than others (with the caveat that AP computation for any class is sensitive to the number of positives for that class in the test set). A similar observation is true for different objects given an interaction. In addition, we observe that interactions which can occur with only a specific object category (as indicated by absence of box) such as ``kick-ball'' and ``flip-skateboard'' are easier to detect than those that tend to occur with more than one object such as ``cut'' and ``clean'' and could have drastically different visual and spatial appearance depending on the object. 

\subsection{Qualitative Results}
Qualitative results (Fig.~\ref{fig:ch4_qual_results}) demonstrate the advantages of building HOI detectors on the strong foundation of object and pose detectors. False positives are more commonly due to incorrect interaction prediction than incorrect object/pose detection. 
Notice that cues for preventing false positives could be as subtle as gaze direction as in the case of ``inspect airplane'' and ``watch bird.'' 

\vspace{-0.2cm}
\section{Conclusion}
\vspace{-0.2cm}
We propose a no-frills approach to HOI detection which is competitive with existing literature despite its simplicity. This is achieved through appropriate factorization of the HOI class probabilities, direct encoding and scoring of layout, 
and improved training techniques. Our ablation study shows the importance of human and object appearance, coarse layout, and fine-grained layout for  HOI detection. We also evaluate the significance of the proposed training techniques which can easily be incorporated into other factored models. 
\\
\noindent\textbf{Acknowledgments:} Supported in part by  NSF 1718221, ONR MURI N00014-16-1-2007, Samsung, and 3M.

{\small
\bibliographystyle{ieee_fullname}
\bibliography{references.bib}
}

\end{document}